%% file: eacl2023.tex
\pdfoutput=1

\documentclass[11pt]{article}

\usepackage{EACL2023}

\usepackage{times}
\usepackage{latexsym}

\usepackage[T1]{fontenc}

\usepackage[utf8]{inputenc}

\usepackage{microtype}

\usepackage{hyperref}
\usepackage{microtype}
\usepackage{hyperref}
\usepackage{booktabs}
\usepackage{multirow}
\usepackage{graphicx}
\usepackage{adjustbox}
\usepackage{amsmath}
\usepackage{amsthm}
\usepackage{amssymb}
\usepackage{amsfonts}
\usepackage{algorithm}
\usepackage{mathtools}
\usepackage{algorithmic}
\usepackage{comment}
\usepackage{subcaption}
\usepackage{enumitem}

\usepackage{bm}
\usepackage{color,soul}

\title{Towards Integration of Discriminability and Robustness for Document-Level Relation Extraction}

\input{authors.tex}

\begin{document}
\maketitle

\input{1-abstract}
\input{2-intro}
\input{3-related}

\input{4-model}

\input{5-experiment}

\input{6-conclu}
\section*{Acknowledgements} 
This research is supported by Singapore Ministry of Education’s AcRF Tier 1 Grant (R-253-000-146-133) to Stanley Kok. Any opinions, findings, conclusions, or recommendations expressed in this material are those of the authors, and do not reflect the views of the funding agencies.

\input{7-limitation}

\bibliographystyle{acl_natbib}
\bibliography{custom}

\input{8-appendix}

\end{document}

%% file: authors.tex
\author{Jia Guo\thanks{$^{*}$This work was partially done when Jia Guo was an intern at DAMO Academy, Alibaba Group.}$^{~1,2}$, Stanley Kok$^{1}$, Lidong Bing$^{2}$ \\
        $^{1}$School of Computing, National University of Singapore \\
        $^{2}$DAMO Academy, Alibaba Group\\
        \texttt{guojia@u.nus.edu, skok@comp.nus.edu.sg} \\
        \texttt{l.bing@alibaba-inc.com}
        }

%% file: 1-abstract.tex
\begin{abstract}
Document-level relation extraction (DocRE) predicts relations for entity pairs that rely on long-range context-dependent reasoning in a document. As a typical multi-label classification problem, DocRE faces the challenge of effectively distinguishing a small set of positive relations from the majority of negative ones. This challenge becomes even more difficult to overcome when there exists a significant number of annotation errors in the dataset. In this work, we aim to achieve better integration of both the discriminability and robustness for the DocRE problem. Specifically, we first design an effective loss function to endow high discriminability to both probabilistic outputs and internal representations. We innovatively customize entropy minimization and supervised contrastive learning for the challenging multi-label and long-tailed learning problems. To ameliorate the impact of label errors, we equipped our method with a novel negative label sampling strategy to strengthen the model robustness. In addition, we introduce two new data regimes to mimic more realistic scenarios with annotation errors and evaluate our sampling strategy. Experimental results verify the effectiveness of each component and show that our method achieves new state-of-the-art results on the DocRED dataset, its recently cleaned version, Re-DocRED, and the proposed data regimes. \footnote{Our codes and datasets are available at \url{https://github.com/guojiapub/PEMSCL}.}
\end{abstract}

%% file: 2-intro.tex
\section{Introduction}

The problem of document-level relation extraction (DocRE) has garnered increasing attention from the research community~\citep{Quirk17, Peng17, DocRED} due to its importance to real-world applications. DocRE is inherently a multi-label problem, in which we have to predict a set of relations from the pre-defined label set for every entity pair in a document. Thus, it is crucial for DocRE models to adopt an effective learning objective that can clearly distinguish massive semantically close relations. 

Recently, several works have proposed new loss functions to learn an adaptive threshold for better separating positive and negative relations. However, these approaches \citep{ATLOP, KD-DocRE} either enforce learning a \textit{total order} among all relations that leads to superfluous comparisons and diminishing differences among them or improperly penalize all pre-defined labels of positive entity pairs if their average margins are lower than the threshold when addressing the label imbalance problem \citep{NCRL}. In contrast, we propose an approach that learns a \textit{partial order}, ranking all positive relations above a threshold individually, which is in turn ranked above all negative relations. Our approach does not waste precious data and probability mass in modeling the ordering among positive relations (likewise for negative relations). We further sharpen the distinction in each distribution of a relation and the threshold through the principled use of entropy minimization.

Besides, none of the above methods take the discriminability of internal representations into account, as well as the model robustness against annotation errors. To solve these issues, we introduce novel modifications to the supervised contrastive learning \citep{SCL} to accentuate the differences among the embeddings of entity pairs from different classes and the similarities of that from the same class. Our method can better accommodate the multi-label setting and the long-tail phenomenon that is typically present in DocRE datasets. To combat the annotation error problem stated in \citet{Re-DocRED}, we design two new data regimes and a novel negative label sampling strategy that gives consistently strong performance even with incomplete annotations.
In sum, our contributions are three-fold: 
\begin{itemize}
     \setlength{\itemsep}{1pt}
    \setlength{\parsep}{1pt}
    \setlength{\parskip}{1pt}
    \item We propose an effective loss function that boosts the discriminability of both internal embeddings and probabilistic outputs.
    \item We achieve good integration of discriminability and robustness by incorporating a novel negative label sampling strategy.
    \item Experimental results consistently demonstrate that we achieve new state-of-the-art performance in a variety of settings.
\end{itemize}

%% file: 3-related.tex
\section{Related Work}
\label{sec:app_related}
\paragraph{Document-level relation extraction (DocRE)} Early works on DocRE focus on utilizing graph convolutional networks (GCNs) \citep{GCN} to conduct complex cross-sentence reasoning on a document graph \citep{GCNN, EoG, GLRE, SIRE}. Recently, methods fine-tuned on large pre-trained language models \citep{BERT, Roberta} achieved significant performance gain. In particular, SSAN \citep{SSAN} encoded entity dependencies into the self-attention mechanism to strengthen context and entity reasoning. ATLOP \citep{ATLOP} employed the multi-head attention weights to generate entity-related context representations which enhanced the embeddings of entity pairs. To better address the multi-label classification problem, both ATLOP \citep{ATLOP} and NCRL \citep{NCRL} proposed to treat the NA class as an adaptive threshold. DocuNet \citep{DocuNet} and KD-DocRE \citep{KD-DocRE} extended the ATLOP architecture by increasing interactions between entities and incorporating knowledge distillation, respectively. Besides, other DocRE models attempted to leverage auxiliary information for relation prediction, such as meta dependency paths \citep{LSR}, external knowledge bases \citep{MIUK}, and evidences \citep{Eider, SAIS}. We additionally provide detailed comparison with existing works in Section~\ref{sec:thresh-loss}.

\paragraph{Other related works} Entropy Minimization technique was commonly seen in semi-supervised learning works \citep{Entro_Mini, Vu19}. However, we are the first to employ entropy minimization in the challenging multi-label supervised learning framework. Besides, our entropy minimization takes effect in each customized probability distribution of the relation label and threshold class, which will encourage a larger distinction between them. 

Supervised contrastive learning (SCL) \citep{SCL} extends self-supervised contrastive learning \citep{MoCo, SimCLR} to the fully supervised setting by constructing ``positive'' and ``negative'' examples based on their labels. ERICA \citep{ERICA} proposed a pre-training framework using contrastive learning to improve representations of entities and relations. However, this work samples positive pairs for relations proportionally to their total amount of examples, which will lead to biased optimization that favors primary relations over minor ones. Besides, they only maximize the similarity of one positive example pair each time, which may weaken the global effect of clustering. Instead, we give equal consideration to each relation and each positive example of anchors, and elaborately tailored the supervised contrastive learning to suit both the multi-label problem and long-tailed relation learning.

%% file: 4-model.tex
\section{Methodology}
In this section, we describe our model called PEMSCL that is based on a \textbf{P}airwise moving-threshold loss,  \textbf{E}ntropy \textbf{M}inimization, and \textbf{S}upervised \textbf{C}ontrastive \textbf{L}earning.

\subsection{Problem Formulation}
Let $D\!=\!\{w_l\}_{l=1}^{L}$ be a document containing $L$ words and a set of entities $\mathcal{E}_D = \{e_i\}_{i=1}^{|\mathcal{E}_D|}$. Each entity $e_i$ is associated with a set of mentions $\mathcal{M}_{e_i}\!=\!\{m^i_j\}_{j=1}^{|\mathcal{M}_{e_i}|}$ (i.e., a set of phrases referring to the same entity $e_i$). In document-level relation extraction, we predict the subset of relations in a predefined set $\mathcal{R}\!=\!\{r_k\}_{k=1}^{|\mathcal{R}|}$ that hold between each pair
of entities $(e_h, e_t)_{h,t=1,\ldots,|\mathcal{E}_D|,h \neq t}$. We sometimes abbreviate an entity pair $(e_h,e_t)$ as $(h,t)$ to simplify notation. A relation is deemed to exist between the head entity $e_h$ and tail entity $e_t$ if it is expressed between any of their corresponding mentions. If no relation exists between any pair of their mentions, the entity pair is labeled {\tt NA}. For each entity pair, we term a relation that holds between its constituent entities as \textit{positive}, and the remaining relations in $\mathcal{R}$ as \textit{negative}. An entity pair that is {\tt NA} does not have any positive relation, and has the entire set $\mathcal{R}$ as negative relations (we could consider such a pair as having a special {\tt NA} relation between them).  Document-level relation extraction can be viewed as a multi-label problem, in which an entity pair corresponds to a training/test \textit{example}, and the relations in $\mathcal{R} \cup \{${\tt NA}$\}$ correspond to the possible \textit{labels} or \textit{classes} of the example.

\subsection{Encoder Model}
\label{sec:encoder}
We leverage ATLOP \citep{ATLOP} as our encoder since recent work~\citep{Eider, NCRL} has borne out its usefulness as a backbone in neural architectures. 
For each entity pair $(e_h, e_t)$, the encoder model generates the entity pair representation $\bm{x}_{h,t} \in \mathbb{R}^{d_x}$, and its unnormalized score vector $\bm{f}_{h,t} \in \mathbb{R}^{|\mathcal{R}|+1}$ for relation prediction, we briefly describe them as follows\footnote{Please refer to Appendix~\ref{sec:atlop-encoder} for the computation details.}:
\begin{align}
    \bm{x}_{h,t} & = \text{Encoder}\big ((e_h, e_t)|D, \mathcal{M}_{e_h}, \mathcal{M}_{e_t}\big) \label{eq:atl_enc}\\
    \bm{f}_{h,t} &= \text{Linear}(\bm{x}_{h,t})
    \label{eq:atl_enc1}
\end{align}

\subsection{Pairwise Moving-Threshold Loss with Entropy Minimization}
\label{sec:thresh-loss}
In document-level relation extraction, a fixed probability threshold (e.g. a hyperparameter tuned on the development dataset) is used to decide the boundary of positive and negative relations. However, such a threshold is only suitable for entity pairs \textit{on average}, and may not be ideal for entity pairs with particular properties. 

To address this problem, we design a loss function that utilizes the {\tt NA} class as a dynamic threshold, learning how best to \textit{move} the threshold in accordance with the regularities present in each entity pair. Specifically, we conduct a \textit{pairwise} comparison between each relation and the {\tt NA} class (separately for each relation), and encourage the prediction scores of each positive relation to be higher than that of the {\tt NA} class, and incentivize the score of the {\tt NA} class to be higher than those of negative relations. In this way, we induce a \textit{partial order} over $\mathcal{R} \cup \{${\tt NA}$\}$ for each entity pair. Note that the positive relations are not compared against each other, and their relative rankings are not modeled (likewise for negative relations). This makes sense in the multi-label setting where we are interested in finding the set of relations that are true without being concerned about their relative degrees of veracity.

Formally, we split the predefined relation set $\mathcal{R} = \mathcal{P}_{h,t} \cup \mathcal{N}_{h,t}$ into two mutually exclusive sets for each entity pair $(h,t)$ in a training set, where $\mathcal{P}_{h,t}$ and $\mathcal{N}_{h,t}$ respectively denote the positive and negative relations of $(h,t)$. As mentioned in Section~\ref{sec:encoder}, we make use of 
$\bm{f}_{h,t} \in \mathbb{R}^{|\mathcal{R}|+1}$ that is computed by Equation~\ref{eq:atl_enc}. We denote the elements of $\bm{f}_{h,t}$ that correspond to relation $r \in \mathcal{R}$ and to the {\tt NA} class as $f_r$ and $f_{\eta}$ respectively. (Both $f_r$ and $f_{\eta}$ represent unnormalized prediction scores (logits).) Using $f_r$ and $f_{\eta}$, we compute the probability that the label $C$ of entity pair $(h,t)$ is $r$ (or $\eta$) \textit{conditioned} on $C$ being either $r$ or $\eta$, i.e., \linebreak $P_{h,t}(C\!=\!r|C\!=\!\{r,$ {\tt NA}$\})$ and $P_{h,t}(C\!=\!\eta|C\!=\!\{r,$ {\tt NA}$\})$ respectively, as follows.

\begin{figure}[t]
\centering
\includegraphics[width=0.5\textwidth]{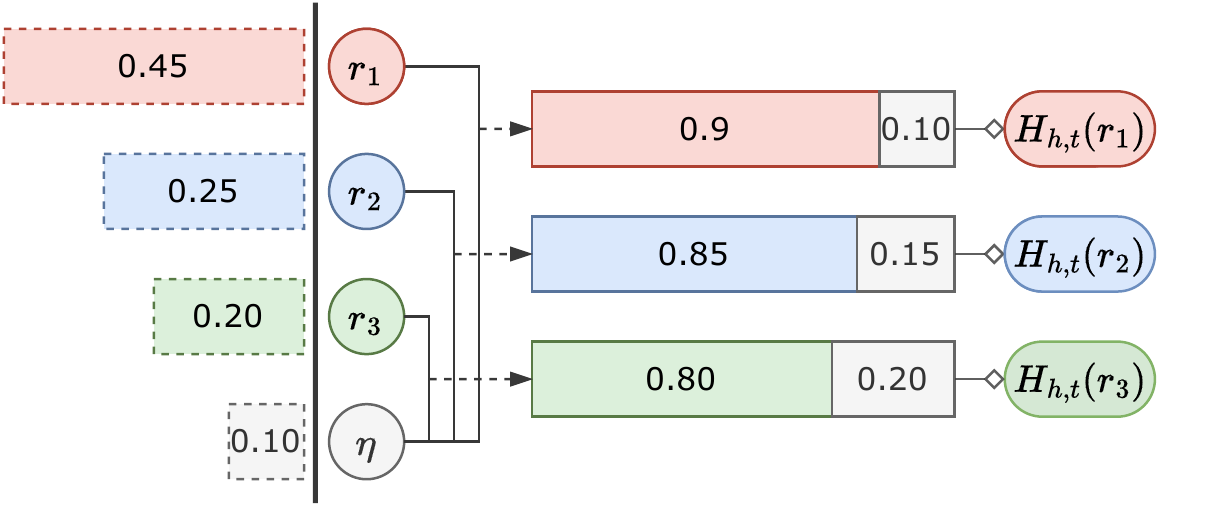} 
\caption{Each positive relation (colored rectangles with solid lines) exhibits a large probability difference from the threshold class (white rectangle with solid lines) when they are separately compared (like what is achieved with our $\mathcal{L}^{h,t}_{pmt}$ loss. We further expand this difference by minimizing $H(r)$ as stated in Eq.~\ref{eq:entro}). However, the probabilities are diminished when each positive relation (colored rectangles with dashed lines) is made to compete with the other, reducing the disparity between the probability of each positive relation and that of the threshold class (white rectangle with dashed lines).}
\label{Fig.loss} 
\end{figure}
\vspace{-1em}
\begin{align}
    P^r_{h,t}(r) & = \frac{\exp(f_r)}{\exp(f_r) + \exp(f_\eta)}, \nonumber \\
    P^{\eta}_{h,t}(r) &\!=\!1\!-\!P^r_{h,t}(r) 
    \!=\! \frac{\exp(f_\eta)}{\exp(f_r)\!+\!\exp(f_\eta)}, \label{eq:p_eta}
\end{align}
where we have abbreviated $P_{h,t}(C\!=\!r|C\!=\!\{r,$ {\tt NA}$\})$ and $P_{h,t}(C\!=\!\eta|C\!=\!\{r,$ {\tt NA}$\})$ as $P^r_{h,t}(r)$ and $P^{\eta}_{h,t}(r)$ respectively.

Our \textit{pairwise moving-threshold} loss $\mathcal{L}^{h,t}_{pmt}$ that maximizing the joint probability of all relations for an entity pair $(h,t)$ is defined as:
\begin{align}
    \mathcal{L}^{h,t}_{pmt} & = - \log \Big (\!\!\prod_{r\in \mathcal{P}_{h,t}}\!\!P^r_{h,t}(r) \!\!\prod_{r \in \mathcal{N}_{h,t}}\!\! \big (1 - P^r_{h,t}(r) \big) \Big ) \nonumber \\
    &= -\!\! \sum_{r\in \mathcal{P}_{h,t}}\!\!\log P^r_{h,t}(r)-\!\! \sum_{r\in \mathcal{N}_{h,t}} \!\!\log P^{\eta}_{h,t}(r) \nonumber \\
    &= \sum_{r\in \mathcal{P}_{h,t}} \log(1 + \exp(f_\eta - f_r)) \nonumber \\ 
    &+ \sum_{r\in \mathcal{N}_{h,t}} \log(1 + \exp(f_r-f_\eta)).
    \label{eq:ptm} 
\end{align}
In Equation~\ref{eq:ptm}, note that the same threshold $f_{\eta}$ is used for all $r \in \mathcal{R}$ for an entity pair $(h,t)$. From the equation, we see that minimizing $\mathcal{L}^{h,t}_{pmt}$ equates to learning scores such that $f_r > f_\eta$ when $r$ is a positive relation, and such that $f_\eta > f_r$ when $r$ is negative relation. The (relative) scores for relations $f_r$ and for the threshold $f_{\eta}$ are fully learned from training data, and are tailored to individual entity pairs. Hence, they can better model the peculiarities specific to each entity pair.

Although previous work \citep{ATLOP, KD-DocRE} employed a similar thresholding mechanism, they learn a \textit{total order} for all relations (or a set of relations) and the threshold class. This wastes finite probability mass (total value of 1.0) in modeling the superfluous ordering among the relations that is not beneficial to multi-label problem, and inevitably diminishes the difference between the probability of each relation and that of the threshold. See Figure~\ref{Fig.loss} for illustration.

Intuitively, a desirable trait of a loss function is that it reduces the uncertainty about whether a relation is positive or negative, thereby allowing its value to be discerned easily. To achieve this in a principled manner, we employ the principle of entropy minimization~\citep{Grandvalet21}. Entropy minimization is typically used on unlabeled data in unsupervised or semi-supervised learning \citep{berthelot2019remixmatch}. In our case, we apply it on \textit{labeled} data in a supervised setting. The information entropy for each pairwise probability distribution between relation $r$ and the threshold class {\tt NA} for entity pair $(h,t)$ is defined as:
\begin{equation}
    \resizebox{.89\linewidth}{!}{$\begin{aligned}
    H_{h,t}(r) = - P^r_{h,t}(r) \log P^r_{h,t}(r) - P^{\eta}_{h,t}(r)\log P^{\eta}_{h,t}(r).  \label{eq:entro}
\end{aligned}$}
\end{equation}
In Equation~\ref{eq:entro}, information entropy decreases as the absolute difference between $P^r_{h,t}(r)$ and $P^{\eta}_{h,t}(r)$ increases, attaining a maximum when $P^r_{h,t}(r) = P^{\eta}_{h,t}(r) = 0.5$ and a minimum when either probability is 1.0 (and the other is 0.0). Thus, incorporating entropy into our loss function would help to accentuate the disparity between the pair $P^{r}_{h,t}(r)$ and $P^{\eta}_{h,t}(r)$ for all relations, making it easier to distinguish a positive (or negative) relation from the threshold {\tt NA}.

We formulate our final pairwise moving-threshold loss with entropy minimization as follows: 
\vspace{-0.2em}
\begin{align}
    \mathcal{L}^{h,t}_{em} &\!=\! \frac{1}{\gamma_1}\!\sum_{r\in \mathcal{P}_{h,t}}\!H_{h,t}(r) \!+ \frac{1}{\gamma_2}\!\sum_{r\in \mathcal{N}_{h,t}}\!H_{h,t}(r), \label{eq:em}\\
    \mathcal{L}_1 &= \sum_{(h, t) \in \mathcal{B}} \mathcal{L}^{h,t}_{pmt} + \mathcal{L}^{h,t}_{em},
    \label{eq:l1}
\end{align}

where $\mathcal{B}$ refers to a training batch, and $\gamma_1 =\{1, |\mathcal{P}_{h,t}|\}$ and $\gamma_2 = \{1,|\mathcal{N}_{h,t}|\}$ are hyperparameters weighting the effect of entropy minimization.

It is noted that using $L_{pmt}$ on its own would lead to poor optimization for positive relations, in the situation where there is a preponderance of negative relations, the sum over $\mathcal{N}_{h,t}$ in Equation~\ref{eq:ptm} might overwhelm the sum over $\mathcal{P}_{h,t}$ to such an extent that pushing $f_{\eta}$ to a large value far above that of $f_r$ for every negative relation $r$ in order to minimize $\mathcal{L}^{h,t}_{pmt}$ (the same issue that also affects previous work \citep{NCRL} without being properly addressed). Instead, our entropy minimization via $\mathcal{L}^{h,t}_{em}$ in Equation~\ref{eq:em} provides a \textit{principled} means to ``balance'' the sharp disparity between the probability of $r$ and that of $\eta$ across \textit{all} relations. Empirically, $\mathcal{L}^{h,t}_{em}$ also demonstrates its efficacy in an ablation study (see Section~\ref{sec:ablation}.)

\subsection{Supervised Contrastive Learning for Multi-Labels and Long-Tailed Relations}
Rather than focusing only on sharpening the disparity of probability outputs as stated in Equation~\ref{eq:l1}, we also seek to accentuate the disparities for the embeddings of entity pairs that are labeled with different relations. To do so, we take inspiration from supervised contrastive learning~\citep{SCL} which aims to  ``pull'' the embeddings of similar examples together, and ``push'' those of dissimilar examples apart. 

However, the original supervised contrastive learning technique only deals with single-label data, and does not handle long-tail distributions. We have to introduce some novel modifications for it to work on our multi-label problem. We make use of the embedding $\bm{x}_{h,t}$ that is computed by Equation~\ref{eq:atl_enc}, and normalized it by L2 normalization before using it in the loss function below. After transplanting the loss function of supervised contrastive learning for our multi-label problem, we obtain the following loss function for an entity pair $(h,t)$: 
\begin{equation}
    \resizebox{.86\linewidth}{!}{$\begin{aligned}
  \mathcal{L}^{h,t}_{scl} = -\log \Big\{ \frac{1}{|\mathcal{S}_{h,t}|}\sum_{p\in \mathcal{S}_{h,t}} \frac{\exp(\bm{x}_{h,t} \cdot \bm{x}_p / \tau)}{\sum\limits_{d \in \mathcal{B},d\neq (h,t)}\exp(\bm{x}_{h,t} \cdot \bm{x}_d/\tau)} \Big\}, \label{eq:scl}
  \end{aligned}$}
\end{equation}

In Equation~\ref{eq:scl}, $\mathcal{B}$ is a batch of examples (entity pairs) including $(h,t)$. $\mathcal{S}_{h,t} \subseteq \mathcal{B}$ is such that each entity pair $p\!=\!(h',t')$ in $\mathcal{S}_{h,t}$ has at least one positive relation in common with $(h,t)$, and $p$ is termed 
a \textit{positive} example of $(h,t)$ (also $(h,t) \notin \mathcal{S}_{h,t}$). The \textit{negative} examples of $(h,t) $ are the remaining examples in the batch, i.e., $\mathcal{B} \setminus (\mathcal{S}_{h,t} \cup \{(h,t)\})$. The operator $\cdot$ refers to the dot product, and $\tau \in \mathbb{R}^+$ is a temperature parameter. To minimize $\mathcal{L}^{h,t}_{scl}$, we maximize the numerator in Equation~\ref{eq:scl} by learning embeddings for $(h,t)$ and its positive examples that are close to each other (according to cosine similarity), and minimize the denominator by learning embeddings for $(h,t)$ and its negative examples that are far apart.

Equation~\ref{eq:scl} would work for document-level relation extraction (DocRE) if not for the long-tail phenomenon that is typically present in DocRE datasets. For example, in the datasets used for our experiments, the top 10 relations account for about 60\% of entity pairs in the dataset. Thus, we often find that an entity pair $(h,t)$ with only long-tailed positive relations does not have any other entity pair in the same batch that has that relation in common, i.e., $|\mathcal{S}_{h,t}| = 0$. This means that Equation~\ref{eq:scl} could not be applied to such entity pairs.
To take such an entity pair $(h,t)$ into account, we design the following loss term:
\begin{equation}
  \mathcal{L}^{h,t}_{lt} =\log \sum\limits_{d \in \mathcal{B},d\neq (h,t)}\exp(\bm{x}_{h,t} \cdot \bm{x}_d / \tau), \label{eq:lt}
\end{equation}

in which we solely maximize the dissimilarities between the embedding of $(h,t)$ and those of other entity pairs in the same batch $\mathcal{B}$.
The final loss function for supervised contrastive learning is:
\begin{equation}
  \mathcal{L}_2\!=\!\!\sum_{(h,t)\in \mathcal{B_P}}\!\!\mathbb{I}_{\{|\mathcal{S}_{h,t}|\neq 0\}} \mathcal{L}^{h,t}_{scl} +\! \mathbb{I}_{\{|\mathcal{S}_{h,t}|= 0\}} \mathcal{L}^{h,t}_{lt}, \label{eq:l2}
\end{equation}
where $\mathbb{I}_{\{\}}$ is an indicator function that takes the value of 1 if the condition in ${\{\}}$ is satisfied, and the value of 0 otherwise.
In Equation~\ref{eq:l2}, $\mathcal{B_P} \subseteq \mathcal{B}$ is a subset of entity pairs in a batch that is labeled with at least one relation in $\mathcal{R}$. In other words, $\mathcal{B_P}$ does not contain any entity pair that is labeled with the {\tt NA} class (i.e., all relations in $\mathcal{R}$ are considered negative for the entity pair), since it does not make sense to minimize the embedding distance between two entity pairs that are labeled {\tt NA}, and thus have no relation in common. 

Combining Equations~\ref{eq:l1} and \ref{eq:l2}, we obtain the final loss function that is used for training our model:
\begin{equation}
  \mathcal{L} =  \mathcal{L}_1 + \lambda\mathcal{L}_2, \label{eq:l}
\end{equation}
where $\lambda \in \mathbb{R}^+$ is a hyperparameter.

\subsection{Negative Label Sampling}
\label{sec:neg-label}
As reported by recent works \citep{DocRED_scratch, Re-DocRED}, the DocRE benchmark suffers from the severe false-negative problem, which means that quite a few entity pairs previously labeled as {\tt NA} class should have at least one relation label. Blithely ignoring this issue will greatly harm the performance of the method and cause ill-defined evaluation. To enhance the robustness of our method, we propose a novel negative label sampling strategy, which only samples a small fraction of negative relations for each entity pair with {\tt NA} label when computing the loss function. We assume that the true relation labels for those false-negative examples are hard to be sampled from the massive negative relations, thus we could avoid erroneously treating the correct labels as negative relations in the loss function. 

Let $\mathcal{B_N} \subseteq \mathcal{B}$ denote the subset of all entity pairs that are labeled {\tt NA} in a current batch $\mathcal{B}$. For each entity pair $(h,t)$ in $\mathcal{B_N}$, we uniformly sample a subset of negative relations $\mathcal{N}'_{h,t} \subseteq \mathcal{N}_{h,t}$, and define the following loss function:
\begin{equation}
    \resizebox{.86\linewidth}{!}{$\begin{aligned}
    \mathcal{L}' =\!\!\!\!\! \sum_{(h, t) \in \mathcal{B_N}} \sum_{r\in \mathcal{N}'_{h,t}} -\log P^{\eta}_{h,t}(r) \! + \! \frac{1}{\gamma_2}\!\!\sum_{r\in \mathcal{N}'_{h,t}}H_{h,t}(r), \label{eq:ns}
    \end{aligned}$}
\end{equation}
where $P^{\eta}_{h,t}(r)$ and $H_{h,t}(r)$ are defined in Equation~\ref{eq:p_eta} and Equation~\ref{eq:entro} respectively.

Let $\mathcal{B_P} = \mathcal{B} \setminus \mathcal{B_N}$ denote the subset of entity pairs in the current batch $\mathcal{B}$ that is labeled with at least one positive relation. Combining terms in Equations~\ref{eq:l1}, \ref{eq:l2}, and \ref{eq:ns}, we obtain the final loss function $\mathcal{L}^{\tt NA}$ that incorporates our sampling approach:
\begin{align}
\mathcal{L}^{\tt NA}_1 &= \mathcal{L}' + \sum_{(h, t) \in \mathcal{B_P}} \mathcal{L}^{h,t}_{pmt} + \mathcal{L}^{h,t}_{em}, \nonumber \\
\mathcal{L}^{\tt NA} &= \mathcal{L}^{\tt NA}_1 + \lambda \mathcal{L}_2.
\label{eq:l_na}
\end{align}

Observe that $\mathcal{L}^{\tt NA}_1$ has modified $\mathcal{L}_1$ (Equation~\ref{eq:l1}) by changing the latter's sum over negative relations for entity pairs that are labeled {\tt NA}. Also note that the loss $\mathcal{L}_2$ due to supervised contrastive learning remains unchanged in Equation~\ref{eq:l_na} because it operates at the level of entities (specifically their embeddings) rather than at the level of relation labels.

Although previous papers~\citep{Empirical_NER, Rethink_NER} seem to adopt a similar negative sampling strategy, our approach has significant differences from them. The previous works sampled negative instances (i.e., entire entity pairs with {\tt NA} labels in our case) and removed those unselected negative instances from the training dataset. In our approach, we sample negative \textit{labels} of {\tt NA} entity pairs, and do not discard any entity pairs, making our approach potentially more data efficient.

%% file: 5-experiment.tex
\section{Experiments}
\label{sec:expt}

\subsection{Benchmark Description}
\label{sec:app_bench}
DocRED \citep{DocRED} is a large-scale dataset constructed from Wikipedia and Wikidata, and is widely used as a benchmark for document-level relation extraction (DocRE). 
However, recent studies~\citep{DocRED_scratch, Re-DocRED} have found that many entity pairs (or examples) that are labeled {\tt NA} are erroneous, and should be instead labeled with at least one positive relation in $\mathcal{R}$. To ameliorate this problem, Re-DocRED~\citep{Re-DocRED} relabels the original training and development sets of DocRED and splits DocRED's development set into two equal halves as new development and test sets, respectively. Instead of comparing models on the faulty DocRED dataset, the results on Re-DocRED should be regarded as a fair comparison.

\subsection{Two New Data Regimes}
To evaluate the models in a more realistic experimental setting in which their resilience to noisy data is carefully tested, we propose two new data regimes, OOG-DocRE and OGG-DocRE, that are based on the above DocRED and Re-DocRED benchmarks. Every ``O'' represents the \textbf{O}riginal labels obtained from the original unclean, noisy DocRED dataset; similarly, every ``G'' represents the \textbf{G}old labels in the new, cleaned Re-DocRED dataset. Each letter in ``OOG'' and ``OGG'' represent different sources of labels for training, validation, and testing, respectively. 
Both regimes reflect the real-world scenario where training data is noisy, and manual effort can only be expended on cleaning a relatively small validation/test set. All models are trained and tuned only on the training and validation sets respectively, and evaluated on the test set. Note that in both regimes the cleaned training set from Re-DocRed is not used. Table~\ref{tab:dataset} contains details about the datasets.

\begin{table}[t!]
\centering
\resizebox{\linewidth}{!}{
  \begin{tabular}{lrrr}
    \toprule
      \multirow{2}{*}{\textbf{Dataset}}  & \multicolumn{1}{c}{\textbf{Train}} & \multicolumn{1}{c}{\textbf{Dev}} & \multicolumn{1}{c}{\textbf{Test}}  \\
      & \#Doc / \#Example &  \#Doc / \#Example &  \#Doc / \#Example \\ 
    \midrule
     DocRED & 3,053 / 1,198,650   &  1,000 / 396,790 & 1,000 / 392,158\\
     Re-DocRED & 3,053 / 1,193,092 &  500 / 193,232 & 500 / 198,670 \\
     \midrule
     \multicolumn{4}{l}{\textit{Our new data regimes}} \\
     OOG-DocRE & 3,053 / 1,198,650  &  500\footnotemark/ 195,682 & 500 / 198,670   \\
     OGG-DocRE & 3,053 / 1,198,650 &  500 / 193,232 & 500 / 198,670 \\
    \bottomrule
  \end{tabular}}
\caption{Dataset statistics. We construct two new data regimes based on the \textbf{O}riginal labels from DocRED and \textbf{G}old labels from Re-DocRED. The total number of predefined relation labels for all datasets is 96 (i.e., |$\mathcal{R}|=96$).}
\label{tab:dataset}
\end{table} 
\vspace{-1mm}
\footnotetext{500 documents share the same titles as the development set of Re-DocRED, but labeled by DocRED.}

\begin{table*}[!t]
\centering
\resizebox{0.8\linewidth}{!}{
\begin{tabular}{lcccc}
\toprule
& \multicolumn{2}{c}{DocRED Dev} & \multicolumn{2}{c}{DocRED Test}  \\
  \cmidrule(r){2-3}  \cmidrule(r){4-5}
  \textbf{Model} & Ign $F_1$ & $F_1$ & Ign $F_1$ & $F_1$ \\
\midrule
\textit{Implemented on DeBERTa$_{\text{Large}}$} & \multicolumn{4}{c}{} \\
ATLOP \citep{ATLOP} & 62.16$\pm$0.15 & 64.01$\pm$0.12 & 62.12 & 64.08 \\
ATLOP + BCE \citep{NCRL} & 61.92$\pm$0.13 & 63.96$\pm$0.15 & 61.83 & 63.92\\
NCRL \citep{NCRL} & \underline{62.98$\pm$0.18} & \underline{64.79$\pm$0.13} & \underline{63.03} & \underline{64.96}\\
\midrule
PEMSCL (Ours) & \textbf{63.25$\pm$0.09} & \textbf{65.15$\pm$0.10} & \textbf{63.40} & \textbf{65.41} \\
\midrule
\midrule
         & \multicolumn{2}{c}{Re-DocRED Dev} & \multicolumn{2}{c}{Re-DocRED Test}  \\
         \cmidrule(r){2-3}  \cmidrule(r){4-5}
        & Ign $F_1$ & $F_1$ & Ign $F_1$ & $F_1$ \\
\midrule
\textit{Implemented on RoBERTa$_{\text{Large}}$} & \multicolumn{4}{c}{} \\
JEREX \citep{JEREX} & 69.12 & 70.33 & 68.97 & 70.25 \\
ATLOP + BCE$^*$ \citep{NCRL} & 75.86$\pm$0.13 & 75.25$\pm$0.11 & 75.91 & 75.36\\
ATLOP \citep{ATLOP} & 76.88 & 77.63 & 76.94 & 77.73\\
DocuNet \citep{DocuNet} & 77.53 & 78.16 & 77.27 & 77.92 \\
KD-DocRE \citep{KD-DocRE} & 77.92 & 78.65 & 77.63 & 78.35 \\
NCRL$^*$ \citep{NCRL} & \underline{78.41$\pm$0.21} & \underline{79.15$\pm$0.20} & \underline{78.45} & \underline{79.19} \\
\midrule
PEMSCL (Ours) & \textbf{79.02$\pm$0.20}  & \textbf{79.89$\pm$0.17}  & \textbf{79.01} & \textbf{79.86}\\
\bottomrule
\end{tabular}}
\caption{Results on DocRED and Re-DocRED. Ign $F_1$ stands for the F1 score excluding relational facts in the training set. Results for baseline models on the test and dev set of DocRED are taken from their original papers. The results on RE-DocRED for NCRL and ATLOP + BCE~\citep{NCRL} (i.e., marked with $^*$) are reproduced by us with their default code\footnotemark $ \ $and our implementation, respectively; other results of baselines on Re-DocRED are taken from~\citep{Re-DocRED}. We report the mean and standard deviation on the development set of 5 runs with different random initialization for our PEMSCL model and the reproduced  baselines, and report the test scores using the best-performing model on the development set. For implementation details, please refer to Appendix \ref{sec:app_imple}.}
\label{tab:main}
\end{table*}

\begin{table}[t!]
\centering
\resizebox{\linewidth}{!}{
  \begin{tabular}{lccccc}
    \toprule
        \textbf{Model} & Dev Ign $F_1$ & Dev $F_1$ & Head $F_1$ & Mid $F_1$ & Tail $F_1$ \\ 
    \midrule
     Ours & \textbf{79.02} & \textbf{79.89} & \textbf{82.99} &	\textbf{75.70} & \textbf{63.51}\\
      -- $\mathcal{L}_{em}^{h,t}$ & 78.38 & 79.17 & 82.35 & 74.75 & 62.35  \\     
      -- $\mathcal{L}_2$ & 78.36 & 79.10 & 82.40 & 74.50 & 62.22 \\
      -- $\mathcal{L}_{em}^{h,t}$ and $\mathcal{L}_2$ & 77.92 & 78.63 & 81.92 & 74.06 & 61.16\\
    \bottomrule
  \end{tabular}}
  \caption{Ablation study of our PEMSCL model on Re-DocRED. ``--'' represents the removal of our model's components. We also report the $F_1$ scores for the top 10 relations (Head $F_1$), the middle 70 relations (Mid $F_1$), and the last 20 relations (Tail $F_1$) ranked by the number of entity pairs that are related by them. The mean result of 3 runs with different random initialization on the development set of Re-DocRED are reported.}
  \label{tab:ablation}
\end{table}

\begin{table*}[t!]
\centering
\resizebox{0.75\linewidth}{!}{
  \begin{tabular}{lcccccc}
    \toprule
       & \multicolumn{2}{c}{Orig-Dev} & \multicolumn{2}{c}{Gold-Dev} & \multicolumn{2}{c}{Gold-Test}  \\
         \cmidrule(r){2-3}  \cmidrule(r){4-5} \cmidrule(r){6-7}
        & Ign $F_1$ & $F_1$ & Ign $F_1$ & $F_1$ & Ign $F_1$ & $F_1$ \\
\midrule
\textit{On OOG-DocRE Regime} &  \multicolumn{4}{c}{} \\
ATLOP \citep{ATLOP} & 60.94 & 62.95 & 46.99 & 47.14 & 47.52 & 47.65
\\
NCRL \citep{NCRL} & 61.42 & 63.52 & 49.06 & 49.21 & 48.41 & 48.53\\
PEMSCL (Ours) & \textbf{62.05} & \textbf{64.19} & \underline{50.82} & \underline{50.99} & \underline{50.92} & \underline{51.10} \\
\midrule
PEMSCL$^\dagger$ (Ours) & 46.07 & 49.51 & \textbf{62.05} & \textbf{63.39} & \textbf{62.76} & \textbf{64.03} \\
\midrule
\midrule
\textit{On OGG-DocRE Regime}  &  \multicolumn{4}{c}{} \\
ATLOP \citep{ATLOP} & - & - & 48.23 & 48.54	& 48.50 & 48.77 \\
NCRL \citep{NCRL} &  - & - & 49.92 & 50.08 & 50.10 &  50.25	\\
PEMSCL (Ours) & - & - &  \underline{50.43} & \underline{50.62} & \underline{51.09} &  \underline{51.25} \\
\midrule
PEMSCL$^\dagger$ (Ours) & - & - & \textbf{62.40} & \textbf{63.72} & \textbf{62.47} & \textbf{63.73}  \\
    \bottomrule
  \end{tabular}}
 \caption{Results on two new data regimes. The best results are \textbf{bolded}, and the second best results are \underline{underlined}. PEMSCL$^\dagger$ refers to our best-performing model on the development set after using our proposed negative label sampling strategy (Section~\ref{sec:neg-label}). The sampling ratio of 0.1 is set with a development set.}
 \label{tab:regime}
\end{table*}

\subsection{Results on DocRE Benchmarks}

From Table.\ref{tab:main}, we see that our PEMSCL model performs the best on both development and test sets of the original DocRED dataset and the cleaned Re-DocRED dataset (the models are trained on their corresponding training sets). It is worth noting that the results among recent models (e.g., DocuNet, KD-DocRE, NCRL) are almost indistinguishable on DocRED (see Appendix~\ref{sec:app_docred}), especially after considering their standard deviations.
However, the performance gaps between models become significant when we validate and test on the Re-DocRED dataset. This strongly suggests that the original DocRED's (overly erroneous) development and test sets  cannot truly ascertain the performance differences between models. In contrast, the cleaned version Re-DocRED provides a more faithful comparison of the models. Henceforth, we analyze model performances based solely on Re-DocRED's development and test sets.

Compared with ATLOP (upon which our model is developed), our PEMSCL model achieves around a 3-point improvement in terms of both Ign $F_1$ and $F_1$ scores on Re-DocRED's development and test sets.  When compared against the recent strong baseline NCRL, our PEMSCL model continues to do better, achieving about a 1-point improvement in terms of $F_1$ score on Re-DocRED's development set. After taking the standard deviations into account, the results still show that PEMSCL outperforms NCRL. In sum, the above results demonstrate the effectiveness of our proposed model, and ascertain that it has achieved new state-of-the-art performances.

\subsection{Ablation Study}
\label{sec:ablation}

In addition to the main metrics $F_1$ and Ign $F_1$, we also report the $F_1$ scores for different types of relations. We first rank in descending order all predefined relations by the number of entity pairs that are labeled with them. Next, we classify them into three categories: \textit{head} relations (the top 10 relations, accounting for 64\% of Re-DocRED's training data), \textit{tail} relations (the bottom 20 relations, accounting for 2\% of training data), and \textit{middle} relations (the remaining relations). 

From Table~\ref{tab:ablation}, we see that each component plays a pivotal role in the effectiveness of our PEMSCL model -- removing a component or a combination of them compromises performance. Removing the $\mathcal{L}_{em}^{h,t}$ and $\mathcal{L}_2$ components individually results in a performance decline of 0.90\% and 0.99\% in terms of $F_1$ score respectively. When either of these two components is removed, we see a sharper decline in terms of Tail $F_1$ (1.82\% and 2.03\%) than in terms of Head $F_1$ (0.77\% and 0.71\%). This shows that both components are useful for long-tailed relations, and highlights the effectiveness of $\mathcal{L}_2$, part of which is designed to cater to long-tailed relations.
\vspace{-0.07em}
After removing both $\mathcal{L}_{em}^{h,t}$ and $\mathcal{L}_2$ together, the performances on Head $F_1$, Mid $F_1$, and Tail $F_1$ all significantly drop by 1.29\%, 2.17\%, and 3.70\% respectively. Even with only one loss term remaining (i,e, $\mathcal{L}_{pmt}^{h,t}$), our EMSCL model still surpasses the baseline ATLOP on Re-DocRED in Table~\ref{tab:main} by 1.3\% on Dev $F_1$, reflecting the usefulness of our pairwise moving-threshold loss. 

\footnotetext{\url{https://github.com/yangzhou12/NCRL}}

\begin{figure}[htb]
\centering
\includegraphics[width=0.5\textwidth]{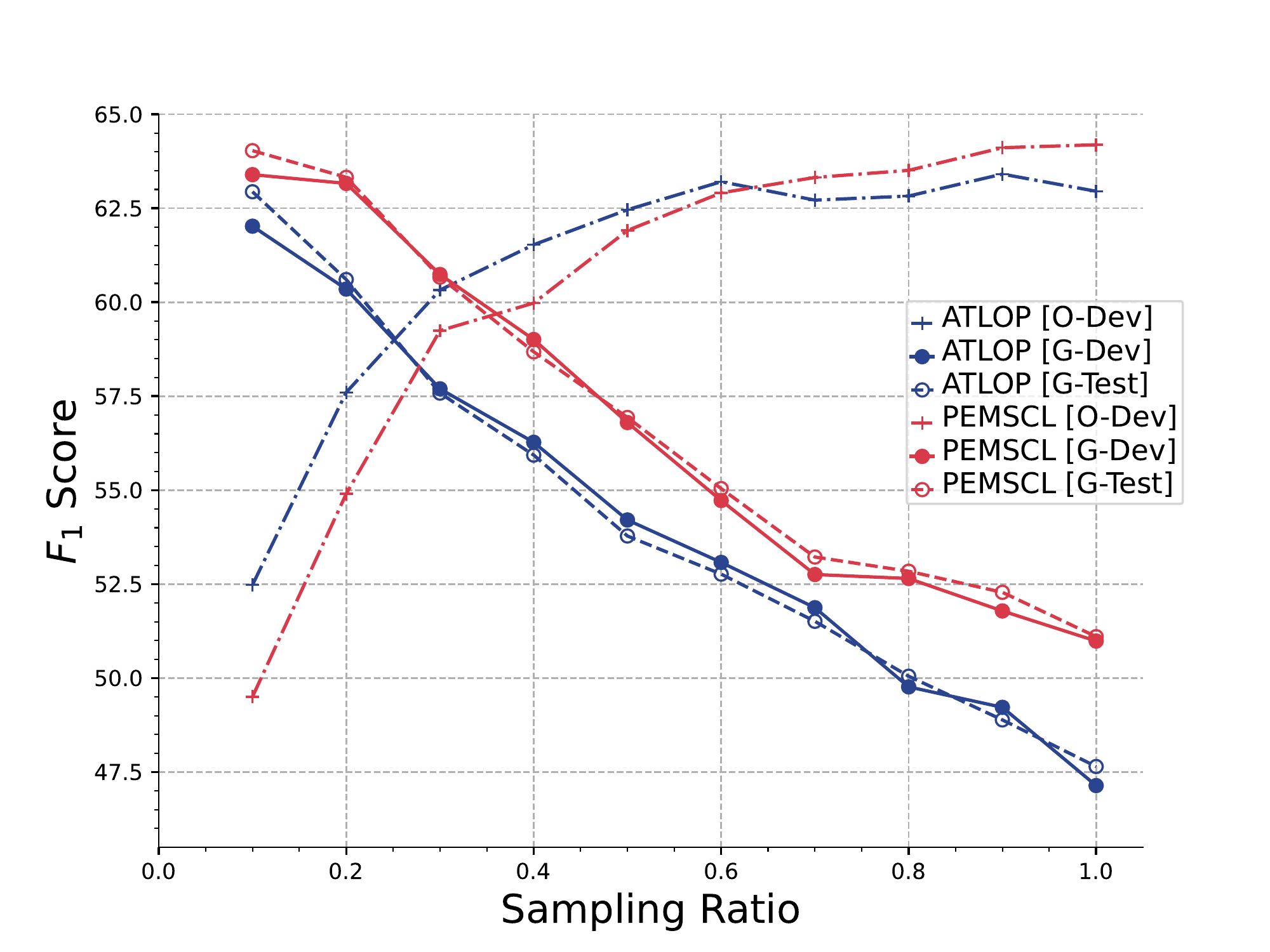} 
\caption{The effect of negative label sampling ratio in the OOG-DocRE regime.}
\label{Fig.oog} 
\end{figure}

\subsection{Results on New Data Regimes}
\label{sec:new_data}
Table~\ref{tab:regime} shows the performance of our PEMSCL model and baselines on our proposed data regimes: OOG-DocRE and OGG-DocRE. We select NCRL as a focal baseline from among the recent baselines due to its competitive performance on Re-DocRED. 
We see that all models perform above 62-point $F_1$ on Orig-Dev (the original development set from DocRED) when trained on the original training dataset.  
However, when we evaluate all models on Gold-Dev and Gold-Test (the clean development and test sets from Re-DocRED), the performances of the models (including ours) dramatically decrease by around 15-point $F_1$ on both Gold-Dev and Gold-Test. Upon inspection, we find that the models misclassify a lot of positive examples (entity pairs) as {\tt NA}, which is an expected outcome of being misguided by the erroneous false-negative labels in DocRED. 

However, after using our proposed negative label sampling loss (Equation~\ref{eq:l_na}), our PEMSCL model exhibits tremendous improvement on both Gold-Dev and Gold-Test by 24\% and 25\% on the $F_1$ scores respectively. 
This demonstrates the effectiveness of our negative label sampling strategy in countering the noise present in entity pairs that are labeled {\tt NA}. The same conclusion can be drawn from the results for the OGG-DocRE regime. Moreover, we notice that both ATLOP and NCRL improve by at least 1-point $F_1$ on the OGG regime compared with the OOG regime. This demonstrates the usefulness of gold labels even in a small amount. However, our model performs comparably on both regimes, indicating the stability of our model in different regimes.  

We also investigate the effect of the sampling ratio on our proposed strategy. The sampling ratio refers to the ratio of negative labels that we keep during the training for each entity pair that is labeled {\tt NA}. We apply our negative label sampling approach on both ATLOP and our PEMSCL model. As seen from Figure ~\ref{Fig.oog}, PEMSCL consistently performs better than ATLOP by a clear margin. We also find that the performances of the models on Gold-Dev and Gold-Test gradually decrease as the sampling ratio is increased. This is because as we keep more (purportedly) negative labels in our loss function, the risk of wrongly penalizing potentially true labels increases concomitantly. 

\begin{figure*}[t]
\centering
\includegraphics[width=\textwidth]{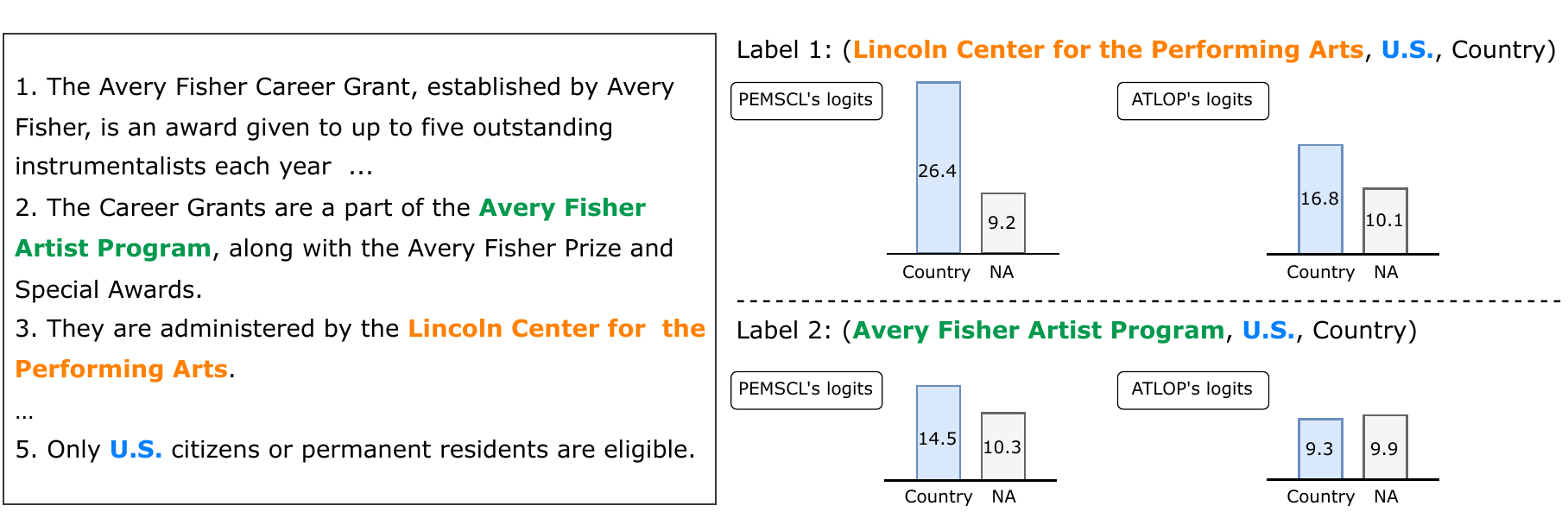}
\caption{Case Study.}
\label{Fig.case} 
\end{figure*}
We observe that the sampling ratio has the opposite effect on Orig-Dev.  As the sampling ratio increases, the $F_1$ on Orig-Dev increases, leading one to mistakenly conclude that a large sampling rate should be used. This provides strong evidence of the poor data quality in DocRED, and shows how it can misguide training and lead to poor results. The results on the OGG-DocRE regime are similar (see Appendix~\ref{sec:app_ogg} for details).

\subsection{Case Study}
Figure~\ref{Fig.case} shows a case study of our proposed PEMSCL model and the baseline ATLOP model. We can see that for the entity pair (\textit{Lincoln Center for the Performing Arts}, \textit{U.S.}), both models successfully detect the correct relation label, i.e., \texttt{Country}. However, the logit difference between the \texttt{Country} relation and the threshold label \texttt{NA} in our model is much larger than that of the ATLOP model (26.4-9.2 > 16.8-10.1). This demonstrates that our model is capable of learning a more differentiated distribution of the final probability scores. For the entity pair of (\textit{Avery Fisher Artist Program}, \textit{U.S.}), the ATLOP model fails to correctly predict its label and classifies it as \texttt{NA} class since the logit of \texttt{Country} is lower than that of the threshold class (9.3 < 9.9). However, our model not only correctly predicts its correct label, but also maximizes the discriminability of the prediction scores (14.5 vs 10.3).

%% file: 6-conclu.tex
\section{Conclusions}
In this paper, we propose a novel method for DocRE problem called PEMSCL, which contains a pairwise moving-threshold loss with entropy minimization, adapted supervised contrastive learning, and a novel negative sampling strategy, to achieve good integration of both discriminability and robustness. Experimental results show that our method achieves new state-of-the-art results.

%% file: 7-limitation.tex
\section*{Limitations}
First, we require a large amount of GPU resources to conduct our experiments because we deal with large document-based datasets (whose input text is significantly longer than those of traditional sentence-level tasks). Second, we implement our model on two large pre-trained language models, Roberta-large~\citep{Roberta} and Deberta-large~\citep{Deberta}, both of which also have a large GPU footprint. Third, the performance of our adapted supervised contrastive learning component is dependent on GPU batch size (a larger batch size allows more contrastive examples to be used to learn better embeddings).

%% file: 8-appendix.tex
\appendix

\begin{table*}[t!]
\centering
\resizebox{0.9\linewidth}{!}{
  \begin{tabular}{lcccccccc}
    \toprule
        \textbf{Dataset} & Batch size & Epoch & Warmup ratio & Learning rate & $\tau$ & $\lambda$ & $\gamma_1$ & $\gamma_2$ \\ 
    \midrule
    DocRED & 4 & 5 & 0.10 & 2e-5 & 2.0 & 2.0 & 1 & 1\\
    Re-DocRED & 4 & 8 & 0.06 & 2e-5 & 0.2 & 0.1 & $|\mathcal{P}_{h,t}|$ & $|\mathcal{N}_{h,t}|$\\
    \bottomrule
  \end{tabular}}
  \caption{Best hyperparameters for benchmarks.}
  \label{tab:para}
\end{table*}

\section{Implementation Details}
\label{sec:app_imple}
We mainly implement our method using the pre-trained RoBERTa-large \citep{Roberta} as the encoder model. Due to limited computational resources, we only use a larger model DeBERTa-large \citep{Deberta} on the DocRED benchmark. We conduct grid search for the temperature parameter $\tau$ and the loss coefficient $\lambda$ (\{0.1, 0.2, 0.5, 1.0, 2.0\}), learning rate (\{1e-5, 2e-5, 3e-5\}), and warmup ratio of optimizer (\{0.02, 0.06, 0.10\}). We implement our model in the PyTorch version of Huggingface Transformers\footnote{\url{https://huggingface.co/}}, and run all experiments on a NVIDIA Quadro RTX 8000 GPU. The best hyperparameters used in our experiments are shown in Table\ref{tab:para}.

\begin{table*}[!t]
\centering
\resizebox{0.75\linewidth}{!}{
\begin{tabular}{lcccc}
\toprule
& \multicolumn{2}{c}{DocRED Dev} & \multicolumn{2}{c}{DocRED Test}  \\
  \cmidrule(r){2-3}  \cmidrule(r){4-5}
  \textbf{Model} & Ign $F_1$ & $F_1$ & Ign $F_1$ & $F_1$ \\
\midrule
\textit{Implemented on RoBERTa$_{\text{Large}}$} & \multicolumn{4}{c}{} \\
Coref \citep{Coref} & 57.35 & 59.43 & 57.90 & 60.25\\
SSAN \citep{SSAN} & 60.25 & 62.08 & 59.47 & 61.42\\
ATLOP \citep{ATLOP} & 61.32$\pm$0.14 & 63.18$\pm$0.19 & 61.39 & 63.40\\
DocuNet \citep{DocuNet} & \underline{62.23$\pm$0.12} & 64.12$\pm$0.14 & \underline{62.39} & \textbf{64.55}\\
KD-DocRE \citep{KD-DocRE} & 62.16$\pm$0.10 & \underline{64.19$\pm$0.16} & \textbf{62.57} & \underline{64.28}\\
NCRL \citep{NCRL} & 62.21$\pm$0.22 & 64.18$\pm$0.20 & 61.94 & 64.14 \\
\midrule
PEMSCL (Ours) & \textbf{62.31$\pm$0.19} & \textbf{64.21$\pm$0.17} & 62.17 & \underline{64.28} \\
\bottomrule
\end{tabular}}
\caption{Experimental results on the DocRED dataset.}
\label{tab:docred}
\end{table*}

\section{The Effect of Sampling Ratio on the OGG-
DocRE Setting}
\label{sec:app_ogg}
\begin{figure}[ht]
\centering
\includegraphics[width=0.45\textwidth]{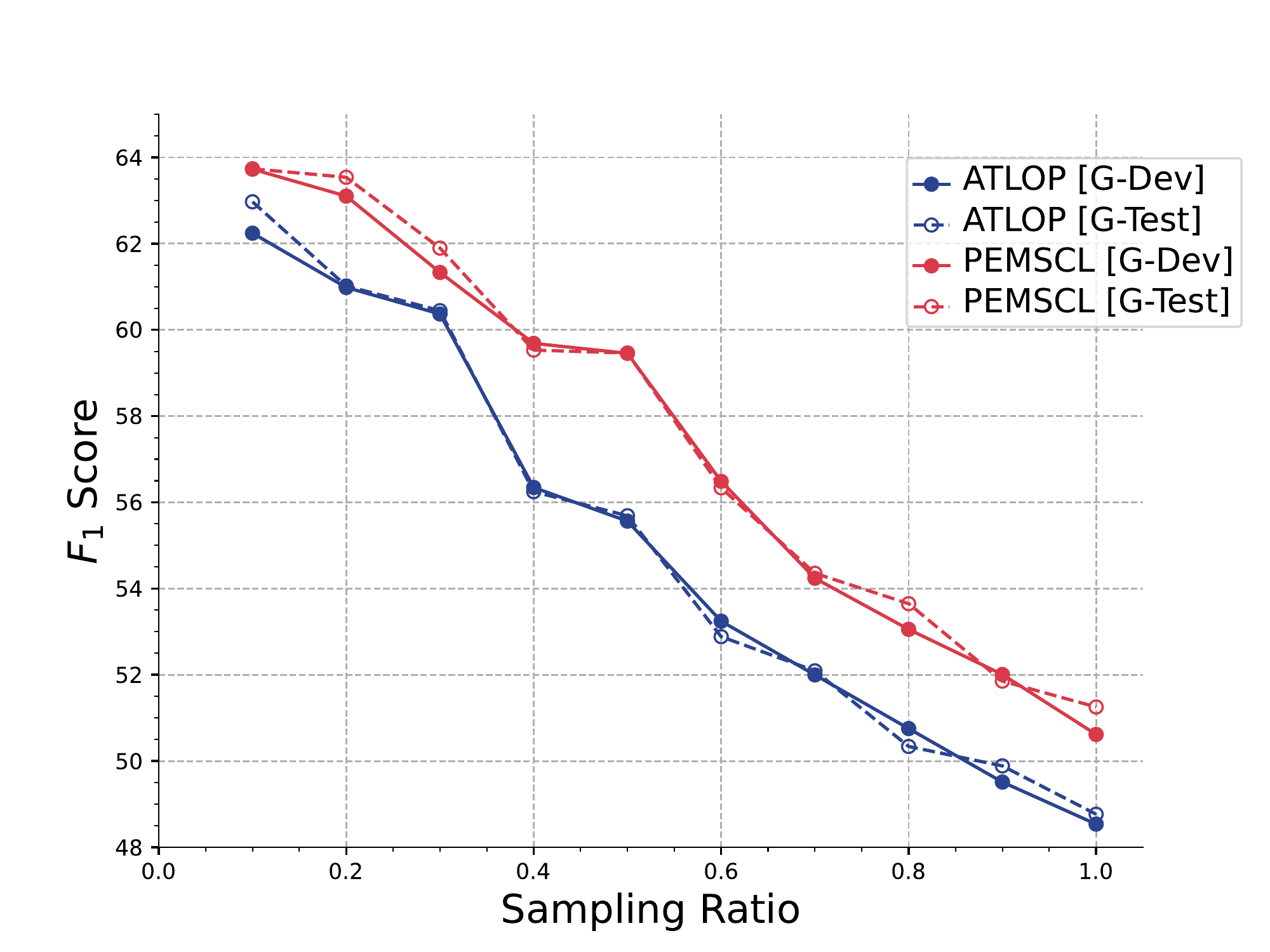} 
\caption{The effect of negative label sampling ratio in the OGG-DocRE regime.}
\label{Fig.ogg} 
\end{figure}

We analyze the effect of the negative label sampling ratio in the OGG-DocRE regime, which is shown in Figure \ref{Fig.ogg}. It presents a similar pattern with that of the OOG-DocRE regime as described in Section \ref{sec:new_data}.

\section{Results on the DocRED Dataset}
\label{sec:app_docred}
We provide the results of RoBERTa-large based models on DocRED in Table~\ref{tab:docred} for a complete comparison. However, these results can not reflect a faithful performance comparison due to the preponderance of erroneous labels in the DocRED dataset. Instead, the results on the Re-DocRED dataset should be taken as a reliable fair comparison.

\input{4-background}

%% file: 4-background.tex
\section{Background: ATLOP Encoder}
\label{sec:atlop-encoder}

For every document, the encoder model first marks each entity mention with a special token ``*'' at its start and end positions, and then feeds the resulting document $D\!=\!\{w_l\}_{l=1}^{L}$ into a pre-trained language model (PLM) to obtain contextual embeddings for each of the document's $L$ tokens: $\mathbf{H}\!=\!\left[\bm{h}_1,\ldots, \bm{h}_L\right] = \text{PLM}(\left[w_1,\ldots, w_L\right])$ where $\bm{h}_l \in \mathbb{R}^d$. ATLOP regards the embedding of ``*'' at the start position of mention $m_j^i$ as its representation, i.e., $\bm{h}_{I(m_j^i)}$, where $I(.)$ is a function mapping a mention $m_j^i$ to the index of its representative ``*'' in $\mathbf{H}$. Next, the embedding $\bm{h}_{e_i} \in \mathbb{R}^d$ of each entity $e_i$ is obtained with logsumexp pooling: 
\vspace{-2mm}
\begin{equation*}
    \bm{h}_{e_i} = \log \sum_{j=1}^{\mathcal{M}_{e_i}} \exp \left( \bm{h}_{I(m_j^i)} \right).
\end{equation*}

For each entity pair $(e_h,e_t)$, ATLOP uses the token-level dependencies present within its multi-head self-attention mechanism to compute a \textit{localized contextual} embedding $\bm{c}_{h,t} \in \mathbb{R}^d$, capturing the contextual information that is relevant to \textit{both} entities $e_h$ and $e_t$. Due to space constraints, we refer readers to \citet{ATLOP} for details on how $\bm{c}_{h,t}$ is computed.

For each entity pair $(e_h, e_t)$, the encoder will generate the final representation $\bm{x}_{h,t}$ for the pair, and its corresponding vector of unnormalized prediction scores $\bm{f}_{h,t} \in \mathbb{R}^{|\mathcal{R}|+1}$ for all relations in $\mathcal{R} \cup \{${\tt NA}$\}$ as follows:
\begin{align}
    [\bm{z}_h^1;\ldots;\bm{z}_h^P] = \bm{z}_h &= \tanh\left(\bm{W}_h \bm{h}_{e_h} + \bm{W}_{c_1} \bm{c}_{h,t}\right), \nonumber \\
    [\bm{z}_t^1;\ldots;\bm{z}_t^P] =\bm{z}_t &= \tanh \left( \bm{W}_t \bm{h}_{e_t} + \bm{W}_{c_2} \bm{c}_{h,t} \right), \nonumber \\
     \bm{x}_{h,t} &= ||_{p=1}^P (\bm{z}_h^p\otimes \bm{z}_t^p), \label{eq:x-ht} \\
  \bm{f}_{h,t} &= \bm{W}_{o}\bm{x}_{h,t} + \bm{b}_o, \label{eq:logit}
\end{align}
where 
$\bm{z}_h, \bm{z}_t \in \mathbb{R}^{d_1}$ are split into $P$ equal-sized groups $[\bm{z}_h^1;\ldots;\bm{z}_h^P]$ and $[\bm{z}_t^1;\ldots;\bm{z}_t^P]$ respectively; $\bm{W}_{\{h,t,c_1,c_2\}} \in \mathbb{R}^{d_1\times d}$, $\bm{W}_o \in \mathbb{R}^{(|\mathcal{R}|+1)\times {d_x}}$, $\bm{x}_{h,t}\in \mathbb{R}^{d_x}$ ($d_x\!=\! \frac{d_1\!\times\!d_1}{P})$, and $\bm{b}_o \in \mathbb{R}^{|\mathcal{R}|+1}$ are \textit{learnable} parameters (in our model too); $\otimes$ is the outer product operator; and the operators $;$ and $||$ respectively represent the concatenation of vectors and matrices.  
The elements in $\bm{f}_{h,t}$ are logits that our model feeds pairwise into (not necessarily the same) softmax functions to obtain relative probabilities between relations (Section~\ref{sec:thresh-loss}).

%% file: eacl2023.bbl
\begin{thebibliography}{34}
\expandafter\ifx\csname natexlab\endcsname\relax\def\natexlab#1{#1}\fi

\bibitem[{Berthelot et~al.(2020)Berthelot, Carlini, Cubuk, Kurakin, Sohn,
  Zhang, and Raffel}]{berthelot2019remixmatch}
David Berthelot, Nicholas Carlini, Ekin~D. Cubuk, Alex Kurakin, Kihyuk Sohn,
  Han Zhang, and Colin Raffel. 2020.
\newblock \href {https://openreview.net/forum?id=HklkeR4KPB} {Remixmatch:
  Semi-supervised learning with distribution alignment and augmentation
  anchoring}.
\newblock In \emph{Proceedings of {ICLR}}.

\bibitem[{Chen et~al.(2020)Chen, Kornblith, Norouzi, and Hinton}]{SimCLR}
Ting Chen, Simon Kornblith, Mohammad Norouzi, and Geoffrey~E. Hinton. 2020.
\newblock \href {http://proceedings.mlr.press/v119/chen20j.html} {A simple
  framework for contrastive learning of visual representations}.
\newblock In \emph{Proceedings of {ICML}}.

\bibitem[{Christopoulou et~al.(2019)Christopoulou, Miwa, and Ananiadou}]{EoG}
Fenia Christopoulou, Makoto Miwa, and Sophia Ananiadou. 2019.
\newblock \href {https://doi.org/10.18653/v1/D19-1498} {Connecting the dots:
  Document-level neural relation extraction with edge-oriented graphs}.
\newblock In \emph{Proceedings of {EMNLP/IJCNLP}}.

\bibitem[{Devlin et~al.(2019)Devlin, Chang, Lee, and Toutanova}]{BERT}
Jacob Devlin, Ming{-}Wei Chang, Kenton Lee, and Kristina Toutanova. 2019.
\newblock \href {https://doi.org/10.18653/v1/n19-1423} {{BERT:} pre-training of
  deep bidirectional transformers for language understanding}.
\newblock In \emph{Proceedings of {NAACL-HLT}}.

\bibitem[{Eberts and Ulges(2021)}]{JEREX}
Markus Eberts and Adrian Ulges. 2021.
\newblock \href {https://doi.org/10.18653/v1/2021.eacl-main.319} {An end-to-end
  model for entity-level relation extraction using multi-instance learning}.
\newblock In \emph{Proceedings of {EACL}}.

\bibitem[{Grandvalet and Bengio(2004)}]{Entro_Mini}
Yves Grandvalet and Yoshua Bengio. 2004.
\newblock \href
  {https://proceedings.neurips.cc/paper/2004/hash/96f2b50b5d3613adf9c27049b2a888c7-Abstract.html}
  {Semi-supervised learning by entropy minimization}.
\newblock In \emph{Proceedings of Neural Information Processing Systems}.

\bibitem[{Grandvalet and Bengio(2005)}]{Grandvalet21}
Yves Grandvalet and Yoshua Bengio. 2005.
\newblock \href
  {https://proceedings.neurips.cc/paper/2004/file/96f2b50b5d3613adf9c27049b2a888c7-Paper.pdf}
  {Semi-supervised learning by entropy minimization}.
\newblock In \emph{Proceedings of Neural Information Processing Systems}.

\bibitem[{He et~al.(2020)He, Fan, Wu, Xie, and Girshick}]{MoCo}
Kaiming He, Haoqi Fan, Yuxin Wu, Saining Xie, and Ross~B. Girshick. 2020.
\newblock \href {https://doi.org/10.1109/CVPR42600.2020.00975} {Momentum
  contrast for unsupervised visual representation learning}.
\newblock In \emph{Proceedings of {CVPR}}.

\bibitem[{He et~al.(2021)He, Liu, Gao, and Chen}]{Deberta}
Pengcheng He, Xiaodong Liu, Jianfeng Gao, and Weizhu Chen. 2021.
\newblock \href {https://openreview.net/forum?id=XPZIaotutsD} {Deberta:
  decoding-enhanced bert with disentangled attention}.
\newblock In \emph{Proceedings of {ICLR}}.

\bibitem[{Huang et~al.(2022)Huang, Hao, Ye, Zhu, Feng, and
  Zhao}]{DocRED_scratch}
Quzhe Huang, Shibo Hao, Yuan Ye, Shengqi Zhu, Yansong Feng, and Dongyan Zhao.
  2022.
\newblock \href {https://aclanthology.org/2022.acl-long.432} {Does
  recommend-revise produce reliable annotations? an analysis on missing
  instances in docred}.
\newblock In \emph{Proceedings of {ACL}}.

\bibitem[{Khosla et~al.(2020)Khosla, Teterwak, Wang, Sarna, Tian, Isola,
  Maschinot, Liu, and Krishnan}]{SCL}
Prannay Khosla, Piotr Teterwak, Chen Wang, Aaron Sarna, Yonglong Tian, Phillip
  Isola, Aaron Maschinot, Ce~Liu, and Dilip Krishnan. 2020.
\newblock \href
  {https://proceedings.neurips.cc/paper/2020/hash/d89a66c7c80a29b1bdbab0f2a1a94af8-Abstract.html}
  {Supervised contrastive learning}.
\newblock In \emph{Proceedings of {NeurIPS}}.

\bibitem[{Kipf and Welling(2017)}]{GCN}
Thomas~N. Kipf and Max Welling. 2017.
\newblock \href {https://openreview.net/forum?id=SJU4ayYgl} {Semi-supervised
  classification with graph convolutional networks}.
\newblock In \emph{Proceedings of {ICLR}}.

\bibitem[{Li et~al.(2021{\natexlab{a}})Li, Ye, Huang, and Zhang}]{MIUK}
Bo~Li, Wei Ye, Canming Huang, and Shikun Zhang. 2021{\natexlab{a}}.
\newblock \href {https://ojs.aaai.org/index.php/AAAI/article/view/17563}
  {Multi-view inference for relation extraction with uncertain knowledge}.
\newblock In \emph{Proceedings of {AAAI}}.

\bibitem[{Li et~al.(2021{\natexlab{b}})Li, Liu, and Shi}]{Empirical_NER}
Yangming Li, Lemao Liu, and Shuming Shi. 2021{\natexlab{b}}.
\newblock \href {https://openreview.net/forum?id=5jRVa89sZk} {Empirical
  analysis of unlabeled entity problem in named entity recognition}.
\newblock In \emph{Proceedings of {ICLR}}.

\bibitem[{Li et~al.(2022)Li, Liu, and Shi}]{Rethink_NER}
Yangming Li, Lemao Liu, and Shuming Shi. 2022.
\newblock \href {https://aclanthology.org/2022.acl-long.497} {Rethinking
  negative sampling for handling missing entity annotations}.
\newblock In \emph{Proceedings of {ACL}}.

\bibitem[{Liu et~al.(2019)Liu, Ott, Goyal, Du, Joshi, Chen, Levy, Lewis,
  Zettlemoyer, and Stoyanov}]{Roberta}
Yinhan Liu, Myle Ott, Naman Goyal, Jingfei Du, Mandar Joshi, Danqi Chen, Omer
  Levy, Mike Lewis, Luke Zettlemoyer, and Veselin Stoyanov. 2019.
\newblock \href {http://arxiv.org/abs/1907.11692} {Roberta: A robustly
  optimized bert pretraining approach}.
\newblock \emph{arXiv preprint arXiv:1907.11692}.

\bibitem[{Nan et~al.(2020)Nan, Guo, Sekulic, and Lu}]{LSR}
Guoshun Nan, Zhijiang Guo, Ivan Sekulic, and Wei Lu. 2020.
\newblock \href {https://doi.org/10.18653/v1/2020.acl-main.141} {Reasoning with
  latent structure refinement for document-level relation extraction}.
\newblock In \emph{Proceedings of {ACL}}.

\bibitem[{Peng et~al.(2017)Peng, Poon, Quirk, Toutanova, and Yih}]{Peng17}
Nanyun Peng, Hoifung Poon, Chris Quirk, Kristina Toutanova, and Wen{-}tau Yih.
  2017.
\newblock \href {https://doi.org/10.1162/tacl\_a\_00049} {Cross-sentence n-ary
  relation extraction with graph lstms}.
\newblock \emph{Trans. Assoc. Comput. Linguistics}, 5:101--115.

\bibitem[{Qin et~al.(2021)Qin, Lin, Takanobu, Liu, Li, Ji, Huang, Sun, and
  Zhou}]{ERICA}
Yujia Qin, Yankai Lin, Ryuichi Takanobu, Zhiyuan Liu, Peng Li, Heng Ji, Minlie
  Huang, Maosong Sun, and Jie Zhou. 2021.
\newblock \href {https://doi.org/10.18653/v1/2021.acl-long.260} {{ERICA:}
  improving entity and relation understanding for pre-trained language models
  via contrastive learning}.
\newblock In \emph{Proceedings of {ACL/IJCNLP}}.

\bibitem[{Quirk and Poon(2017)}]{Quirk17}
Chris Quirk and Hoifung Poon. 2017.
\newblock \href {https://doi.org/10.18653/v1/e17-1110} {Distant supervision for
  relation extraction beyond the sentence boundary}.
\newblock In \emph{Proceedings of {EACL}}.

\bibitem[{Sahu et~al.(2019)Sahu, Christopoulou, Miwa, and Ananiadou}]{GCNN}
Sunil~Kumar Sahu, Fenia Christopoulou, Makoto Miwa, and Sophia Ananiadou. 2019.
\newblock \href {https://doi.org/10.18653/v1/p19-1423} {Inter-sentence relation
  extraction with document-level graph convolutional neural network}.
\newblock In \emph{Proceedings of {ACL}}.

\bibitem[{Tan et~al.(2022{\natexlab{a}})Tan, He, Bing, and Ng}]{KD-DocRE}
Qingyu Tan, Ruidan He, Lidong Bing, and Hwee~Tou Ng. 2022{\natexlab{a}}.
\newblock \href {https://aclanthology.org/2022.findings-acl.132}
  {Document-level relation extraction with adaptive focal loss and knowledge
  distillation}.
\newblock In \emph{Findings of {ACL}}.

\bibitem[{Tan et~al.(2022{\natexlab{b}})Tan, Xu, Bing, Ng, and
  Aljunied}]{Re-DocRED}
Qingyu Tan, Lu~Xu, Lidong Bing, Hwee~Tou Ng, and Sharifah~Mahani Aljunied.
  2022{\natexlab{b}}.
\newblock \href {https://aclanthology.org/2022.emnlp-main.580} {Revisiting
  docred - addressing the false negative problem in relation extraction}.
\newblock In \emph{Proceedings of {EMNLP}}.

\bibitem[{Vu et~al.(2019)Vu, Jain, Bucher, Cord, and P{\'{e}}rez}]{Vu19}
Tuan{-}Hung Vu, Himalaya Jain, Maxime Bucher, Matthieu Cord, and Patrick
  P{\'{e}}rez. 2019.
\newblock \href
  {http://openaccess.thecvf.com/content\_CVPR\_2019/html/Vu\_ADVENT\_Adversarial\_Entropy\_Minimization\_for\_Domain\_Adaptation\_in\_Semantic\_Segmentation\_CVPR\_2019\_paper.html}
  {{ADVENT:} adversarial entropy minimization for domain adaptation in semantic
  segmentation}.
\newblock In \emph{Proceedings of {CVPR}}.

\bibitem[{Wang et~al.(2020)Wang, Hu, Cao, and Sun}]{GLRE}
Difeng Wang, Wei Hu, Ermei Cao, and Weijian Sun. 2020.
\newblock \href {https://doi.org/10.18653/v1/2020.emnlp-main.303}
  {Global-to-local neural networks for document-level relation extraction}.
\newblock In \emph{Proceedings of {EMNLP}}.

\bibitem[{Xiao et~al.(2022)Xiao, Zhang, Mao, Yang, and Han}]{SAIS}
Yuxin Xiao, Zecheng Zhang, Yuning Mao, Carl Yang, and Jiawei Han. 2022.
\newblock \href {https://arxiv.org/abs/2109.12093} {{SAIS:} supervising and
  augmenting intermediate steps for document-level relation extraction}.
\newblock In \emph{Proceedings of {NAACL}}.

\bibitem[{Xie et~al.(2022)Xie, Shen, Li, Mao, and Han}]{Eider}
Yiqing Xie, Jiaming Shen, Sha Li, Yuning Mao, and Jiawei Han. 2022.
\newblock \href {https://aclanthology.org/2022.findings-acl.23} {Eider:
  Empowering document-level relation extraction with efficient evidence
  extraction and inference-stage fusion}.
\newblock In \emph{Findings of {ACL}}.

\bibitem[{Xu et~al.(2021)Xu, Wang, Lyu, Zhu, and Mao}]{SSAN}
Benfeng Xu, Quan Wang, Yajuan Lyu, Yong Zhu, and Zhendong Mao. 2021.
\newblock \href {https://ojs.aaai.org/index.php/AAAI/article/view/17665}
  {Entity structure within and throughout: Modeling mention dependencies for
  document-level relation extraction}.
\newblock In \emph{Proceedings of {AAAI}}.

\bibitem[{Yao et~al.(2019)Yao, Ye, Li, Han, Lin, Liu, Liu, Huang, Zhou, and
  Sun}]{DocRED}
Yuan Yao, Deming Ye, Peng Li, Xu~Han, Yankai Lin, Zhenghao Liu, Zhiyuan Liu,
  Lixin Huang, Jie Zhou, and Maosong Sun. 2019.
\newblock \href {https://doi.org/10.18653/v1/p19-1074} {Docred: {A} large-scale
  document-level relation extraction dataset}.
\newblock In \emph{Proceedings of {ACL}}.

\bibitem[{Ye et~al.(2020)Ye, Lin, Du, Liu, Li, Sun, and Liu}]{Coref}
Deming Ye, Yankai Lin, Jiaju Du, Zhenghao Liu, Peng Li, Maosong Sun, and
  Zhiyuan Liu. 2020.
\newblock \href {https://doi.org/10.18653/v1/2020.emnlp-main.582}
  {Coreferential reasoning learning for language representation}.
\newblock In \emph{Proceedings of {EMNLP}}.

\bibitem[{Zeng et~al.(2021)Zeng, Wu, and Chang}]{SIRE}
Shuang Zeng, Yuting Wu, and Baobao Chang. 2021.
\newblock \href {https://doi.org/10.18653/v1/2021.findings-acl.47} {{SIRE:}
  separate intra- and inter-sentential reasoning for document-level relation
  extraction}.
\newblock In \emph{Findings of {ACL/IJCNLP}}.

\bibitem[{Zhang et~al.(2021)Zhang, Chen, Xie, Deng, Tan, Chen, Huang, Si, and
  Chen}]{DocuNet}
Ningyu Zhang, Xiang Chen, Xin Xie, Shumin Deng, Chuanqi Tan, Mosha Chen, Fei
  Huang, Luo Si, and Huajun Chen. 2021.
\newblock \href {https://doi.org/10.24963/ijcai.2021/551} {Document-level
  relation extraction as semantic segmentation}.
\newblock In \emph{Proceedings of {IJCAI}}.

\bibitem[{Zhou et~al.(2021)Zhou, Huang, Ma, and Huang}]{ATLOP}
Wenxuan Zhou, Kevin Huang, Tengyu Ma, and Jing Huang. 2021.
\newblock \href {https://ojs.aaai.org/index.php/AAAI/article/view/17717}
  {Document-level relation extraction with adaptive thresholding and localized
  context pooling}.
\newblock In \emph{Proceedings of {AAAI}}.

\bibitem[{Zhou and Lee(2022)}]{NCRL}
Yang Zhou and Wee~Sun Lee. 2022.
\newblock \href {https://doi.org/10.48550/arXiv.2205.00476} {None class ranking
  loss for document-level relation extraction}.
\newblock In \emph{Proceedings of IJCAI}.

\end{thebibliography}
